\pdfoutput=1

\documentclass[11pt]{article}

\usepackage{acl}

\newcommand{\minus}{\scalebox{0.7}[1.0]{$-$}}

\usepackage{times}
\usepackage{latexsym}

\usepackage[T1]{fontenc}

\usepackage[utf8]{inputenc}

\usepackage{microtype}

\usepackage{xcolor}
\usepackage{times}
\usepackage{latexsym}
\usepackage{microtype}
\usepackage{graphicx}
\usepackage{subfigure}
\usepackage{booktabs} 
\usepackage{natbib}
\usepackage{graphicx}
\usepackage{amsmath}
\usepackage{amscd}
\usepackage{amssymb}
\usepackage{amsthm}
\usepackage{mathabx}
\usepackage{algorithmic}
\usepackage{algorithm}
\usepackage{booktabs}
\usepackage{multirow}

\usepackage{paralist}
\usepackage{color, colortbl}
\usepackage{enumitem}
\usepackage[normalem]{ulem}
\usepackage{stackengine}

\title{Using natural language prompts for machine translation}

\author{Xavier Garcia \\
  \texttt{xgarcia@google.com} \\\And
  Orhan Firat \\
  \texttt{orhanf@google.com} \\}

\begin{document}
\maketitle
\begin{abstract}
We explore the use of natural language prompts for controlling various aspects of the outputs generated by machine translation models. We demonstrate that natural language prompts allow us to influence properties like formality or specific dialect of the output. We show that using language names to control the output language of multilingual translation models enables positive transfer for unseen language pairs. This unlocks the ability to translate into languages not seen during fine-tuning by using their English names. We investigate how scale, number of pre-training steps, number of languages in fine-tuning, and language similarity affect this phenomenon.
\end{abstract}

\section{Introduction}

Neural machine translation (MT) has become the defacto approach for building high-fidelity translation models for high-resource language pairs. To mitigate the difficulties in training neural MT models for low-resource language pairs, researchers have turned to \emph{multilingual machine translation} \cite{DBLP:journals/corr/FiratCB16,johnson2017google} as an alternative. In this setting, we use data from all the available language pairs and train a single model on this combined corpus. To grant users control over the output language, practitioners include language-specific parameters in the model architecture. The most popular strategy is the original one proposed by \citet{johnson2017google}: include an additional token in the input, typically denoted as \texttt{<2xx>} \textit{tokens} or  \textit{language tags} \cite{wu2021language}. For example, to translate the phrase ``How are you?'' to Spanish, we would prepend the text with the special symbol \texttt{<2es>} as follows: $$\textnormal{How are you?} \rightarrow \texttt{<2es>}\textnormal{ How are you?}$$ During training, the model will learn to associate the task of translation into Spanish with the token associated to the string \texttt{<2es>}, allowing the user to translate any input into Spanish by prepending \texttt{<2es>} onto the input before passing it to the model. This strategy boasts a variety of benefits: it requires no modification to the underlying translation model architecture and only incurs minor memory costs per language. 

Multilingual MT as a paradigm has been quite successful due to the presence of \textit{positive transfer}: joint-training on both low and high-resource language pairs leads to a disproportionate boost in performance for the low-resource language pairs \cite{arivazhagan2019massively}. This emergent phenomenon is a recurring theme in the contemporary literature of natural language processing. Large models trained on unsupervised objectives on massive text corpora can be finetuned on new tasks and typically perform better than models trained from scratch \cite{devlin2018bert,radford2018improving,conneau2019unsupervised}. Moreover, learning from multiple tasks has been shown to be helpful for generalization at all stages of training: during pre-training  \cite{aribandi2021ext5}, as a special training stage before fine-tuning \cite{aghajanyan-etal-2021-muppet}, and even during fine-tuning \cite{tang2020multilingual}. In all cases, one attains stronger positive transfer by training on larger number of tasks, intensifying the need for a flexible way to handle an arbitrary number of tasks. 

In an effort to obviate the need for artificial task-specific tokens, researchers have begun to explore the use of \emph{natural language prompts} as a way to bias the model. The crux of the technique lies in formulating all tasks in a unified \textit{text-to-text} format \cite{raffel2019exploring}, which requires including an additional text instructions as part of the text input. Re-using our previous example, to translate the sentence "How are you?" to Spanish, we would replace the input with the text "Translate to Spanish: How are you?" Such formatting has many advantages: most notably, it allows us to include any arbitrary task into our models by prepending a task description onto the input. The most fascinating aspect of this method is that it allows for models to potentially tackle tasks unseen during training. Large language models can generalize to these unseen instructions, obtaining reasonable performance in a wide variety of tasks.  Moreover, recent works \cite{wei2021finetuned,sanh2021multitask} have shown that we can improve the performance of this instruction-following behavior by fine-tuning on a multi-task mixture using natural language descriptions of the tasks, which mirrors closely the results from the multilingual MT literature.

In this work, we investigate the use of natural language prompts in the context of machine translation. We first demonstrate that machine translation models respond positively to natural language prompts by showing that we can influence the formality of translation outputs produced by black-box machine translation models. Given the success of this initial experiment, we investigate the use of natural language prompts as an alternative to language tags. More precisely, we propose to adapt multilingual models to use prompts of the form \texttt{Translate to Spanish:} instead of the corresponding language tag \texttt{<2es>} through a short fine-tuning phase using parallel data. This approach enables the possibility of performing translations into specific dialects and even unseen languages during fine-tuning by using their corresponding language names. This technique does not introduce any additional parameters nor labeled data, allowing us to potentially deal with arbitrary languages, locales, registers, etc.  We investigate possible factors which impact performance on this unseen language translation task, such as amount of pre-training data, model size, number of language pairs and language similarity. Moreover, we demonstrate that adaptating multilingual translation models with this approach improves performance on zero-shot language pairs.

\section{Related works}

\paragraph{Text generation through prompting} Previous work \cite{kumar2016ask} used a more general formulation of prompting where instead of prepending text, we instead consider an input \emph{template} that contains a slot for the input. In this formulation, we could formulate our prompts as \texttt{Translate to \{language\_name\}: \{input\_slot\}}. \citet{wei2021finetuned} and \citet{sanh2021multitask} have shown this ability to follow natural language instructions can be improved by finetuning the model on a diverse mixture of tasks. All of these works have typically focused on large, English-centric language models. The space of prompting multilingual models, on the other hand, remains underexplored, only having recently been studied by \citet{lin2021few}.

\paragraph{Language bias for multilingual machine translation} Previous work have also explored alternatives to language tags, for example using language embeddings that we add directly to the intermediate representations of the decoder \cite{DBLP:journals/corr/FiratCB16} or subword embeddings \cite{wang2018three,conneau2019unsupervised}, task-specific attention \cite{blackwood2018multilingual}, language tags on the decoder side \cite{liu2020multilingual}, etc. The idea of prompts of the form \texttt{Translate to \{language\_name\}:} was first introduced by \citet{raffel2019exploring}, but they did not compare this strategy with language tags, nor did they explore the effects of this strategy on unseen tasks. 

\section{Preliminary exploration} \label{sec:formality}

\begin{figure*}
\centering
\includegraphics[scale=0.3]{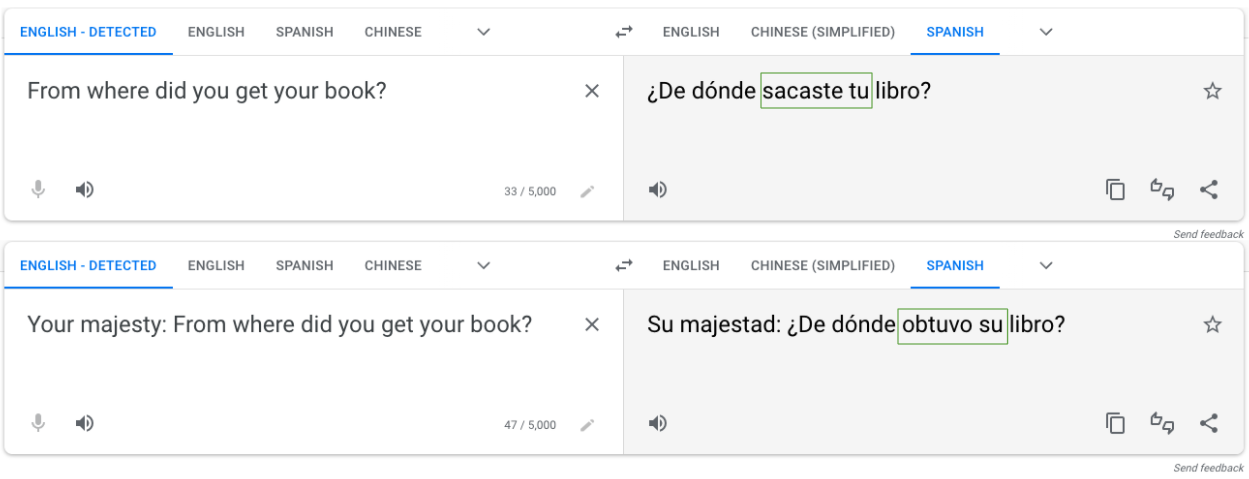}
\caption{\textbf{Using prompts to control for T\minus{}V distinction in Google Translate.} We notice that without the prompt, the model uses the informal second person pronoun \textit{tu}. After prompting, it is easy to extract the desired translation which uses the formal pronoun. We have enclosed the prompt and its translation in red, and the use of the second person pronoun in green. Note that the inclusion of the prompt not only made the model output the formal second person pronoun, but also resulted in the model using the word \emph{obtuvo} (obtain), which is subjectively more formal than \emph{sacaste} (take out). We obtained these results from the API on Tuesday, February 1, 2021.}
\label{fig:api}
\end{figure*}

In this section, we evaluate the feasibility of controlling formality in translation outputs using prompts. We focus on controlling the use of the grammatical T\minus{}V distinction \cite{brown-gilman-1960} which distinguishes between informal and formal second person pronouns. To fully showcase the power of natural language prompts, we use a black-box translation model by using the freely-available Google Translate\footnote{https://translate.google.com/}, which only allows us to specify input language, target language, and the text to be translated. 

We use the Newscrawl dataset\footnote{We use the files from the year 2019, found in https://data.statmt.org/news-crawl/en.} and extract 100 English sentences containing the second person pronoun ``you''.\footnote{More precisely, we filter out sentences which do not contain the substring `` You '' or `` you ''.} We then consider translating these 100 sentences into French, Spanish, German, Telugu and Turkish and keep track of whether the output contains a second person pronoun and whether it is the informal or formal one. We are solely interested on whether we can encourage this black-box model to use the formal second person pronoun. To accomplish this task, we prepend every text input with the phrase "Your majesty:" before producing the translation, then removing the translated prompt.\footnote{In practice, we do this by splitting on the colon. In particular, for a given output \texttt{translation}, we perform the following operation in Python: \texttt{':'.join(translation.split(':')[1:])}} We show an explicit example for the English-Spanish language pair in Figure \ref{fig:api}. To assess which second person pronoun is used, we ask native speakers of each language to inspect the translations and to annotate the usage of the T\minus{}V distinction. To make a formality score, we assign the value of 1 to all instances using the formal second pronoun, \minus{}1 to the informal second person pronoun, and 0 when there is no second person pronoun then add these values up. We report these scores for both the prompted and unprompted inputs in Table \ref{tab:formality}. We note that across all languages, we obtain higher formality scores using the inputs with the prompt than those without.

\section{Experiments} \label{sec:exp}

In this section, we outline a set of experiments to investigate the effects of natural language prompts on the outputs of machine translation models. We first study the effects of using prompts when fine-tuning multilingual models on parallel data. We will show that this finetuning strategy enables multilingual models to translate into languages lacking parallel data with any other language. We then perform an ablation study to understand how language similarity and number of language pairs affects this newfound ability. Finally, we consider adapting a bonafide multilingual translation model with prompts and show that this adaptation improves the performance for language pairs without parallel data with little to no regression on the language pairs seen by the model.

\begin{table}
\centering
\small
\begin{tabular}{cccccccll}
\toprule
Prompt & \textit{fr} & \textit{es} & \textit{de} & \textit{te} & \textit{tr} \\ \midrule 
``Your majesty:'' & 11 & 89 & 74 & 99 & 59   \\
No prompt & \minus{}6 & \minus{}34 & 40 & 93 & 23 \\ \bottomrule
\end{tabular}
  \caption{\textbf{Formality scores for a variety of English-XX language pairs.} Here we define the formality score to be the difference in the number of times the model uses the formal second person pronoun versus informal second person pronoun.}
  \label{tab:formality}
\end{table}

\subsection{Experiments with mT5} \label{sec:mt5}

Given the success of using prompts to control local aspects of a translation, we now consider whether we can use similar prompts to control global aspects: namely, the language of the output itself. 

\begin{table*}
\centering
\small
\begin{tabular}{cccccccll}
\toprule
Models & \# of parameters & \stackanchor{\emph{newstest14}}{\emph{en}$\rightarrow$\emph{fr}} & \stackanchor{\emph{newstest16}}{\emph{en}$\rightarrow$\emph{de}}  & \stackanchor{\emph{newstest16}}{\emph{en}$\rightarrow$\emph{ro}} \\ \midrule 
\emph{Unadapted XXL} & 12.9B & 0.1 & 0.0 & 0.0 \\
\emph{Adapted Large} & 1.23B &21.1 & 15.9 & 15.6    \\
\emph{Adapted XL} & 3.74B & 29.3  & 25.3  & 23.9 \\
\emph{Adapted XXL (10\%)} & 12.9B &  30.9 & 25.2 & 21.0 \\
\emph{Adapted XXL} & 12.9B & \textbf{36.5} & \textbf{32.2}  & \textbf{26.3} \\ \hline
FLAN Zero-shot & 127B & 34.0 & 27.0 & 18.4 \\
GPT-3 Few-shot & 175B & 32.6 & 29.7 & 21.0 \\
XGLM Few-shot & 7.5B & 28.5 & 23.5 & \minus{} \\
Supervised  & Varies & \underline{45.6} & \underline{41.2} & \underline{33.4} \\ \bottomrule
\end{tabular}
  \caption{\textbf{BLEU scores of various models on the traditional WMT language pairs.} We report our models in italics, using the naming convention used in mT5. We refer to the XXL model with no adaptation as \textit{Unadapted XXL}. The rest of our models have been adapted with prompts as described in Section \ref{sec:mt5} and have been analogously labelled with the prefix \textit{Adapted}. The model configuration \textit{Adapted XXL (10\%)} refers to fine-tuning a checkpoint with the same as XXL, but that has only been pre-trained for 102k steps. GPT-3 Few-Shot uses some number of examples between 1 and 64 examples, while XGLM uses 32 examples for each language pair. We mark in bold the best numbers for models that were not trained to perform translation on these language pairs. We underline the best numbers, regardless of model or data avilable. }
  \label{tab:flan_comp}
\end{table*}

\paragraph{Model configurations} We conduct all our experiments in JAX \cite{jax2018github}, using the T5X framework\footnote{https://github.com/google-research/t5x}. For our initial set of experiments, we will be using checkpoints from the mT5 \cite{xue-etal-2021-mt5} family of models as initializations. These are large, multilingual models that have been trained on monolingual data coming from 108 languages. We primarily focus on the XL and XXL configurations (consisting of 3 billion and 13 billion parameters respectively), but also include the Large configuration (consisting of 1 billion parameters) to investigate how the performance changes with scale. We re-use the same tokenizer as mT5, and use a maximum sequence length of 200 tokens, discarding all sequences longer than that during training and truncating the input during inference.

\paragraph{Datasets} We will consider a total of 51 English-centric language pairs.\footnote{Language codes in alphabetical order: af, ar, az, bg, ca, cs, cy, da, el, eo, es, et, eu, fa, fi, ga, gl, hu, id, is, it, ja, ka, km, ko, ku, lt, lv, mg, mk, ms, mt, nl, no, pl, pt, ru, sk, sl, sq, sr, sv, tg, th, tr, uk, ur, uz, vi, xh, zh.} We gather this data from WMT, the OPUS 100 \cite{zhang2020improving} dataset, Paracrawl\footnote{We use the v9 version from https://paracrawl.eu/} and Samanantar \cite{ramesh2021samanantar}. We will draw our monolingual data from the mC4 dataset that was originally used to train the mT5 models.

\paragraph{Data preprocessing} We apply the CLD3 language identification model\footnote{https://github.com/google/cld3} on the mC4 data, and discard any input with less than 95\% probability of being the correct language. As explained in the introduction, we depart from the traditional multilingual MT literature and dispose of the use of language tags. Instead, we will prepend the input with natural language prompts depending on the type of data and language. For monolingual data, we will prepend the input with the text $$\texttt{Infill in \{language\_name\}:}$$ where \texttt{language\_name} corresponds to the English-language name for the corresponding language. We use the word \texttt{Infill} since we use a mask-infilling training objective for this data. For the parallel data, we instead use the prompt $$\texttt{Translate to \{language\_name\}:}$$ where in this setting \texttt{language\_name} corresponds to the English-language name for the target language. At inference, we continue this pattern, even for the languages not seen during fine-tuning. 

\paragraph{Training configurations} Our data sampling strategy consists in first randomly selecting a data source (either monolingual or parallel data) with equal probability, then sampling uniformly from the datasets in the selected source. We continue training with the same learning rate schedule and Adafactor optimizer \cite{shazeer2018adafactor} states as mT5. For parallel data, we use the standard cross-entropy objective. For monolingual data, we use the MASS \cite{song2019mass} objective. We additionally use a dropout rate of 0.1 and label smoothing set to 0.1. 

\paragraph{Evaluation} We use beam search with a beam size of 4 and length penalty of $\alpha = 0.6$ for decoding. We follow the usual convention for evaluating machine translation models by reporting detokenized BLEU scores \cite{papineni2002bleu} using the sacreBLEU \cite{post2018call} library.\footnote{BLEU + case.mixed + numrefs.1 + smooth.exp + tok.13a}

In our first experiment, we include monolingual data from mC4 for all available languages except for the Latinized languages, French, German, Kazakh, Romanian, Nepali, Sinhala and Yoruba so that they remain completely unseen during fine-tuning. We will fine-tune each available mT5 configurations on the mixture for 2000 steps and evaluate every 200 steps. To assess the role of scale and amount of pre-training data, we fine-tune the Large, XL, and XXL configurations of mT5, as well as a checkpoint of mT5-XXL which has pre-trained for only 102k steps, roughly equivalent to 10\% of the total amount of steps taken during the pre-training stage of the traditional mT5 models.\footnote{To ensure fair comparison, we requested such a checkpoint from the original authors.}

\begin{figure*}
\centering
\includegraphics[scale=0.175]{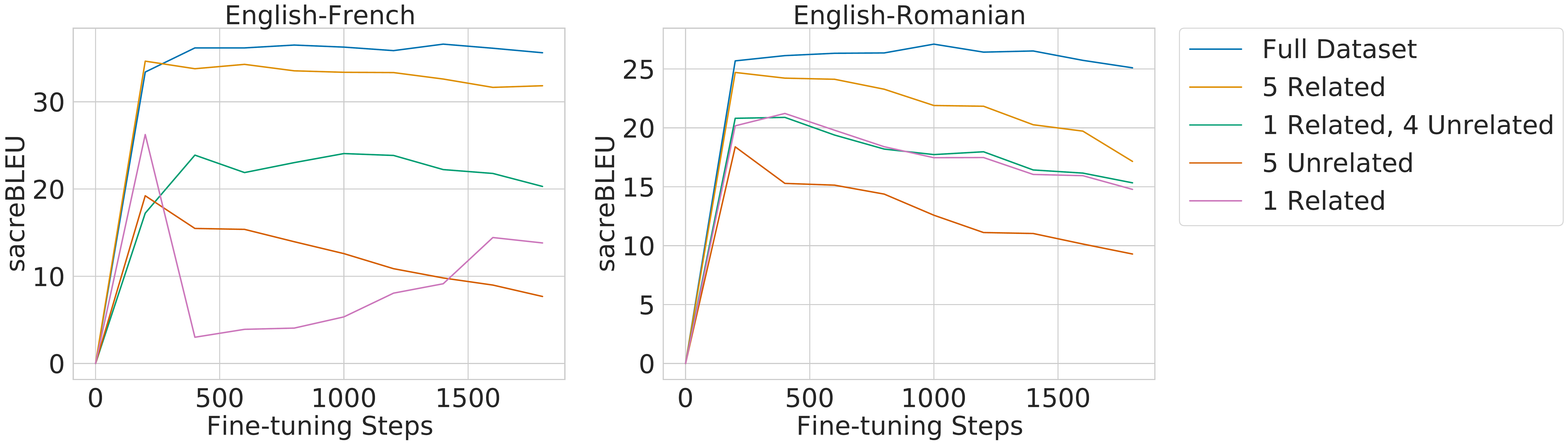}
\caption{\textbf{Influence of language similarity and multi-task setups on unseen language translation performance.} We report on both French and Romanian to control for amount of data seen in pre-training. We note that in the low-resource setting (Romanian), all models show degradation, with only the model leveraging the full dataset of 51 language pairs showing some robustness.}
\label{fig:ablation}
\end{figure*}

We compare our models with other models using natural language: the English-centric GPT-3 \cite{brown2020language} and FLAN \cite{wei2021finetuned}, both of which are many times larger than our models and the multilingual XGLM \cite{lin2021few}. We also consider the current state-of-the-art supervised baselines to serve as reference.  

We first examine the performance across the three language pairs \textit{English-French}, \textit{English-German} and \textit{English-Romanian}, as they are the standard language pairs studied by previous work. In particular, we use the results of \citet{edunov2018understanding} for English-French, the results of \citet{wang2019multi} for English-German, and the results of \citet{caswell2019tagged} for English-Romanian.\footnote{FLAN and GPT-3 compare with the results of \citet{liu2020multilingual} for English-Romanian. However, those results are not using detokenized BLEU, which makes it difficult to compare to their BLUE score, since neither FLAN nor GPT-3 report using the same pre-processing as \citet{liu2020multilingual}. We use the numbers in \citet{caswell2019tagged} since they explicitly use sacreBLEU and report similar BLEU score when using the same setup as \citet{liu2020multilingual}} We report the results in Table \ref{tab:flan_comp}. 

We first note that mT5-XXL is unable to perform translation through prompts without any adaptation, unlike traditional causal decoder-only language models. Next, we note that our largest model outperforms both FLAN and GPT-3 despite being an order of magnitude smaller than both models. This remains true even when we allow GPT-3 to leverage few-shot exemplars. Our models also outperform the comparable multilingual baseline XGLM, which is surprisingly also outperformed by the English-centric models.  Finally, we note the importance of scale: the XXL model which was only pre-trained for a tenth of the time (denoted in Table \ref{tab:flan_comp} as \textit{Adapted XXL (10\%)}) as performs comparably with the XL model on the English-French and English-German.

We test the limits of this approach by evaluating our models on additional language pairs representing different data scenarios: \textit{English-Kazakh}, to examine a low-resource language scenario;\footnote{Kazakh has less monolingual data in mC4 compared to French by two orders of magnitude, and was sampled ten times less often during the pre-training of mT5.} \textit{German-French}, to study the setting where both the source and target languages are unseen during finetuning; \textit{German-Czech}, to investigate the situation where the target language is seen during fine-tuning but the source remains unseen. We include both unsupervised baselines and supervised baselines consisting of winning WMT submissions for the relevant language pairs. In particular, we report the results of \citet{casas2019talp} for English-Kazakh, the results of \citet{xia2019microsoft} for German-French and \citet{marie2019nict} for German-Czech.\footnote{Note that the WMT baseline for German-Czech does not leverage any parallel data at all, which likely explains why our largest model is able to surpass it. The other baselines leverage parallel data for the language pairs in question.}  We report the results in Table \ref{tab:aux_comp}.

\begin{table}
\centering
\small
\begin{tabular}{cccccccll}
\toprule
Models & \stackanchor{\emph{newstest19}}{\emph{en}$\rightarrow$\emph{kk}} & \stackanchor{\emph{newstest19}}{\emph{de}$\rightarrow$\emph{fr}}  & \stackanchor{\emph{newstest19}}{\emph{de}$\rightarrow$\emph{cs}} \\ \midrule 
\emph{Large} & 0.6 & 14.6 & 8.8    \\
\emph{XL} & 5.5  & 16.0 & 16.4 \\
\emph{XXL (10\%)} & 2.5 & 3.4 & 16.0 \\
\emph{XXL} & 7.5 & 20.4 & \textbf{20.6} \\ \hline
WMT  & \textbf{15.5} & \textbf{31.5} & 20.1 \\ \bottomrule
\end{tabular}
  \caption{\textbf{BLEU scores of various models on the additional language pairs.} These language pairs were selected to test for effects on the low-resource setting (English-Kazakh), the setting where both source and target languages are unseen in fine-tuning (German-French) and the case where only the source is unseen (German-Czech). We mark in bold the best numbers for each language pair.}
  \label{tab:aux_comp}
\end{table}

While we could trade training steps for scale in the high-resource languages, such tactics were not as efficient for English-Kazakh or non-English centric pairs. For the case of the German-French in particular, we find the model does actually attain some initial strong performance, but it quickly deteriorates much faster than the English-centric language pairs. In all cases, however, we find that the ability to translate into unseen language begins to degrade overtime, suggesting that the model begins to specialize to the languages seen during fine-tuning and starts to lose its generalizability.

\paragraph{Zero-shot dialect-aware machine translation} Finally, we examine the possibility of translating into specific dialects of languages. This evaluation tests whether our model can not only translate into unseen languages, but also test whether we can apply local changes to the output in the same languages, similar to the experiments considered Section \ref{sec:formality}. For this particular evaluation, we use the a subset of the  TED Talks dataset \cite{qi2018and}. We test our models on the language pairs \emph{English - French Canadian}, \emph{English - Brazilian Portuguese}, \emph{English - Mainland Chinese}, and \emph{English - Taiwanense}. Note that since these language pairs are dialects, we can always obtain some initial performance by translating into the base language similar to our previous experiments. For example, we could use the prompt $$\texttt{Translate to Portuguese:}$$ when evaluating on Brazilian Portuguese. We compare this baseline with the alternative of using a prompt which uses the name of the dialect. Re-using our example of Brazilian Portuguese, we could instead use the prompt $$\texttt{Translate to Brazilian Portuguese:}$$ in order to directly target the dialect in question. This allows us to attempt the task of translating into a dialect of a language without the need for explicitly-labeled data in that dialect i.e. \emph{zero-shot dialect-aware translation}. We report the results of these experiments on the XXL configuration in Table \ref{tab:dialects}. We note that for all dialects, using the dialect name itself results in higher than or equal to the original BLEU scores, with half of them being statistically significant greater at a 0.01 level.\footnote{We follow \citet{koehn2004statistical} and use the method of \emph{paired bootstrap resampling}. We use 100 bootstrapped datasets, as recommended in the original work.} This showcases the potential of adapting the output of translation models to specifics locales at inference without requiring labeled-data in those locales.

\begin{table}
\centering
\small
\begin{tabular}{ccccll}
\toprule
Models & \stackanchor{\emph{TED}}{\emph{fr-ca}} & \stackanchor{\emph{TED}}{\emph{pt-br}}  & \stackanchor{\emph{TED}}{\emph{zh-cn}} & \stackanchor{\emph{TED}}{\emph{zh-tw}} \\ \midrule 
\emph{Language name} & 31.6 & 38.7 & 20.7 & 7.56 \\
\emph{Dialect name} & $32.0^{\dagger}$ & 38.8 & 20.7 & $7.81^{\dagger}$ \\ \bottomrule
\end{tabular}
  \caption{\textbf{BLEU scores of English-XX language pairs on the TED Talks datasets for speecific dialects.} These language pairs were selected to test the capability of enforcing the output to be in a given dialect through natural language prompts, without explicitly requiring any dialect-labeled data. Numbers marked with the superscript $\dagger$ are statistically significantly larger than the alternative at a significance level of 0.01. Note that for both Chinese dialects, we used the \texttt{zh} tokenizer option available in sacreBLEU. }
  \label{tab:dialects}
\end{table}

\subsection{Assessing the value of language similarity and multi-task setups} \label{subsec:sim}

Given the strong performance of our models, we sought to investigate the root cause of this phenomenon. We considered two factors which have been shown to influence the performance of multilingual MT models: language similarity and number of language pairs. We performed four additional ablations by fine-tuning mT5-XXL on four different collections of languages: \begin{enumerate} \item \emph{5 Related}: Spanish, Portuguese, Catalan, Italian and Galician. \item \emph{1 Related, 4 Unrelated:} Italian, Chinese, Korean, Vietnamese, and Thai. \item \emph{5-Unrelated:} Japanese, Chinese, Korean, Vietnamese, and Thai. \item \emph{1 Related:} Italian. 
\end{enumerate}For these experiments, we restrict our use of monolingual data to only the languages with available parallel data. We look at the performance on English-French and English-Romanian. We evaluate every 200 steps and plot the results for language pairs in Figure \ref{fig:ablation}. We point out three notable observations:\begin{enumerate} \item Low-resource languages (Romanian) seem more susceptible to catastrophic forgetting than high-resource ones (French). \item Language similarity is critical for successful unseen language translation, but it is not sufficient to only have one similar language pair. \item Catastrophic forgetting is highly mitigated by the presence of an overabundance of language pairs. \end{enumerate}

\subsection{Adapting multilingual translation models}



All of our previous experiments relied on using the parameters of various mT5 configurations as our initial starting point in training. This choice made it easier to study the problem of translating into languages that lacked language tags since the model had already built some implicit representations for these languages. However, this strategy also has its drawbacks: as demonstrated in Section \ref{subsec:sim}, the model's performance on the unseen language pairs begins to deteriorate early in training, before the model has achieved optimal performance on the language pairs with parallel data. By the time the training procedure has converged, the quality for translation into languages which do not appear in fine-tuning will have vanished.

One attempt at mitigating this would be to start with a model which has already seen all these parallel data and thus has already attained optimal quality. We thus depart from using mT5 and instead consider using a multilingual translation model as our starting point. Since all languages of interest have language tags, the problem of unseen language translation is not well-posed. Instead, we investigate whether the use of natural language prompts can improve the performance of our multilingual translation model across language pairs without parallel data. More precisely, we consider the following two questions: \begin{enumerate} \item What is the impact of using natural language when training multilingual translation models? \item Do we obtain any translation quality improvements on language pairs without parallel data by adapting translation models to use natural language prompts instead of language tags in the same way we adapted the mT5 models? \end{enumerate}




\paragraph{Experimental setup} We consider a setup with 11 languages. We include parallel data for the language pairs English-\{Bengali, Danish, Dutch, French, Hindi, Polish, Spanish\} and include monolingual data for those 8 languages, as well as German, Czech and Gujarati. We use a batch size of 2048 with a maximum sequence length of 200, discarding any sequence longer than that during training and truncating longer sequences during inference. 

As a baseline, we train a multilingual translation model with language tags, dubbed \emph{Language Tags}. We consider two approaches for using natural language prompts when training multilingual translation models: \emph{From Scratch}, by which we mean replacing the language tags with the analogous prompt all throughout training and  \emph{Adaptation}, in which we take the \emph{Language Tags} model and perform a gentle fine-tuning adaptation procedure using prompts instead of language tags, as we did in our mT5 experiments.

We train our translation model for 100k steps with the same training routine as in our initial experiment (with the exception of the languages available). For the \emph{Adaptation} model,  we only adapt the model for 100 steps with the same training hyperparameters and data, with the only difference consisting of the language tags being replaced with natural language prompts as decribed in Section \ref{sec:mt5}. We choose this small number of steps to avoid regression in the previously-seen language pairs (as we saw in our earlier experiments in Section \ref{sec:mt5}), as well as to ensure the gains found are from the natural language prompts rather than continued training.



\begin{table*}
\centering
\small
\begin{tabular}{llccccccccccc}
\toprule
Models & \multicolumn{2}{l}{\stackanchor{\emph{newstest14}}{\emph{en}$\leftrightarrow$\emph{fr}}} & \multicolumn{2}{l}{ \stackanchor{\emph{newstest13}}{\emph{fr}$\leftrightarrow$\emph{es}}}  & \multicolumn{2}{l}{\stackanchor{\emph{newstest16}}{\emph{en}$\leftrightarrow$\emph{de}}} & \multicolumn{2}{l}{\stackanchor{\emph{newstest19}}{\emph{en}$\leftrightarrow$\emph{gu}}} & \multicolumn{2}{l}{\stackanchor{\emph{newstest13}}{\emph{cs}$\leftrightarrow$\emph{de}}}\\ \midrule 
\emph{Language Tags} & 40.0  & \textbf{43.8}  & 18.6 & 14.9 & 23.8 & 24.4 & 23.6 & 11.6 & 9.1 & 4.1 \\
\emph{From Scratch} & \textbf{40.5}  & 43.6  & 14.8 & 15.2 & 25.4 & 4.8 & \textbf{24.0} & \textbf{13.1} & 2.5 & 2.3 \\
\emph{Adapted} & 40.4 & 43.3  & \textbf{34.8} & \textbf{29.2} & \textbf{36.6} & \textbf{25.3} & \textbf{24.0} & 12.5 & \textbf{10.7} & \textbf{13.3} \\ \bottomrule
\end{tabular}
  \caption{\textbf{BLEU scores for multilingual translation models.} The \emph{From Scratch} model uses natural language prompts all through training. The \emph{Language Tags} model uses language tags all throughout training. Finally, the \emph{Adapted} model fine-tune the \emph{Language Tags} model for 100 steps using the same data.  In this setting, we have the following data constraints: English-French has parallel data; both French and Spanish have parallel data with English; German, Czech, and Gujarati have no parallel data at all, with English or otherwise. We mark the best numbers for each language pair in bold.}
  \label{tab:mt_adapt}
\end{table*}

\paragraph{Evaluation} To study this problem, we consider language pairs across four different levels of supervision. For a given language pair $\mathcal{X}\leftrightarrow\mathcal{Y}$, it must satisfy one of the following constraints: \begin{enumerate} \item  $\mathcal{X}\leftrightarrow\mathcal{Y}$ has parallel data. This has been the focus of traditional machine translation literature \cite{bahdanau2014neural,sutskever2014sequence}. \item Both $\mathcal{X}$ and $\mathcal{Y}$ have parallel data with a third language $\mathcal{Z}$ but not with each other. This constraint is generally referred as the \textit{zero-shot} \cite{arivazhagan2019missing,sestorain2018zero, johnson2017google} or \textit{zero-resource} \cite{DBLP:journals/corr/FiratCB16} setting. \item Only one of either $\mathcal{X}$ or $\mathcal{Y}$ possess no parallel data with any language. These language pairs have been the focus of the recent literature on \textit{multilingual unsupervised machine translation} \cite{wang2020cross,li2020reference,garcia-etal-2020-multilingual,garcia-etal-2021-harnessing,chen2021towards}\item Neither $\mathcal{X}$ nor $\mathcal{Y}$ possess any parallel data with any language.  \end{enumerate} 
Given the data available for this experiment, the language pair English$\leftrightarrow$French satisfies the first constraint; French$\leftrightarrow$Spanish satisfies the second one; English$\leftrightarrow$German and English$\leftrightarrow$Gujarati satisfy the third one; and German$\leftrightarrow$Czech satisfies the last one.

\paragraph{Discussion of results} We present the results of three models considered in Table \ref{tab:mt_adapt}. We note a few patterns. On one hand, training with prompts from scratch did not seem to yield much improvement over the model with language tags. In fact, we see large regressions in performance for the language pairs English$\rightarrow$German and Czech$\leftrightarrow$German when compared to the models trained with language tags. We conjecture that with excessive training, the model learns to treat the prompts as multi-token language tags, which only exacerbates the issues that manifest with traditional language tags. Moreover, since the prompts themselves are several tokens long, this also reduces the amount of examples we can pack into a given batch. However, when we instead adapt the model trained with language tags to use prompts, we do see massive improvements in the zero-shot directions, as well as languages without parallel data. Moreover, we see very little change in the performance of the supervised directions for the \emph{Adapted} model, demonstrating that these large gains come at almost no cost. 


\section{Conclusion} In this work, we undertook an initial exploration on using natural language to control the output of translation models. We demonstrated that by prepending certain phrases, one could increase occurence of formal pronouns, as per the T\minus{}V distinction. When finetuning multilingual models for translation tasks, using prompts with English-language names for the languages allows the user to translate to languages seen by the model during pre-training but not during fine-tuning. While our initial approach at training translation models from scratch with this strategy did not yield strong results, we found positive results when adapting already-trained translation models to use prompts through a short adaptation stage.

\bibliography{anthology,custom}
\bibliographystyle{acl_natbib}

\appendix

\section{Example Appendix}
\label{sec:appendix}

This is an appendix.

\end{document}